\documentclass[runningheads]{llncs}
\usepackage[T1]{fontenc}
\usepackage{graphicx,verbatim}
\usepackage{amsmath}
\usepackage{amssymb}
\usepackage[normalem]{ulem}
\usepackage{hyperref}
\usepackage{multirow}
\usepackage{xcolor}
\usepackage[dvipsnames]{xcolor}
\newcommand{\figin}[2][width=1.25cm]{%
    \raisebox{-0.4\height}{\includegraphics[#1]{#2}}%
}

\begin{document}
\title{tFUSOperator: Operator Learning for Transcranial Focused Ultrasound Digital Twins}
\titlerunning{tFUSOperator}
%

\author{Minjee Seo\inst{1} \and
Haris Ghafoor\inst{1} \and
Minju Seol\inst{1} \and
Seonaeng Cho\inst{1} \and
Kyungho Yoon\inst{1,2}\thanks{Corresponding author}}
\authorrunning{M. Seo et al.}
%
\institute{School of Mathematics and Computing (Computational Science and Engineering), Yonsei University, Seoul, Republic of Korea\\
\email{yoonkh@yonsei.ac.kr}\and
Innovative \& Intelligent Computational Science Institute (IN2CSI), Seoul, Republic of Korea}
  
\maketitle              
\begin{abstract}
Transcranial focused ultrasound (tFUS) requires accurate estimation of the intracranial acoustic field, which is distorted by skull-induced aberrations. Numerical solvers are accurate but computationally expensive for digital twins, where the field must be re-estimated repeatedly as treatment conditions change. Existing deep-learning surrogates are fast but typically use voxel-to-voxel regression on a fixed grid, with no mechanism reflecting how acoustic energy propagates through the skull. We instead cast tFUS simulation as an operator learning problem and propose \textit{tFUSOperator}, a coordinate-aware neural operator that maps the free-field pressure, skull anatomy, and treatment parameters to the intracranial field within a shared physical coordinate frame. To our knowledge, this is the first operator-based formulation of tFUS field prediction. On both seen and unseen skulls, the model localizes the acoustic focus accurately—reaching about 90\% and 72\% Dice, respectively—and it performs nearly as well from magnetic resonance (MR) as from computed tomography (CT) input while running $5.6 \times 10^4$ times faster than numerical simulation. These results suggest a fast, radiation-free route to safe and practical digital twins for patient-specific tFUS treatment. The code is available at: \href{https://github.com/CMME-Lab/tFUSOperator.git}{https://github.com/CMME-Lab/tFUSOperator.git}.


\end{abstract}
\section{Introduction}
Transcranial focused ultrasound (tFUS) is a non-invasive technology that delivers acoustic energy to targeted brain regions through the intact skull \cite{10.3389/fneur.2019.00549,Lee2016,DARMANI202251}. In particular, low-intensity focused ultrasound (LIFU) has recently emerged as a reversible, nonablative modality, with growing applications in neuromodulation \cite{Yaakub2025}, blood–brain barrier opening \cite{00004691}, and targeted drug delivery \cite{Haroon2023}. However, the skull introduces acoustic aberrations that distort the intended focal distribution, making accurate estimation of the intracranial acoustic field essential for reliable treatment planning and guidance \cite{Kyriakou01022014,1516024}.

This need is especially acute in digital twin systems for tFUS, where a patient-specific virtual model must continuously reflect the intracranial acoustic field as treatment conditions change \cite{10.1007/978-3-032-07694-6_6}. Such systems require acoustic field estimation that is both accurate and fast enough for interactive, repeated updates during operation. Numerical solvers such as k-Wave \cite{10.1117/1.3360308} remain the gold standard for this estimation, but their substantial computational cost makes them impractical for the real-time, many-query setting required by a digital twin \cite{10500878,10.1007/978}.

To bridge this gap, recent studies have explored deep learning-based surrogate models that predict intracranial pressure fields from anatomical and acoustic inputs in real time \cite{srivastav2025,10.1007/978-3-032-07694-6_6,10.1002/mp.70259,SEO2024108458,JANG2025111157}. While effective at accelerating inference, these approaches share two structural limitations. First, they are formulated as voxel-to-voxel regression on a fixed simulation grid: the mapping is tied to a particular discretization and learns input–output correlations rather than the propagation process that produces the field. Second, the physics of wave propagation enters only implicitly through the training data, as these models lack an architectural mechanism that reflects how acoustic energy propagates and is reshaped by the skull. Combined with transducer settings that are often restricted to a narrow range of configurations, these limitations leave field formation modeled as statistical fitting rather than as a structured physical operator.

In this work, we cast transcranial ultrasound simulation as an operator learning problem and propose \textit{tFUSOperator}, a neural operator that provides a physically structured representation of field formation through coordinate-aware attention over acoustic and anatomical inputs. Our model represents acoustic free-fields, skull images, and treatment parameters—including transducer position and orientation—within a shared physical coordinate frame, allowing these heterogeneous inputs to be jointly attended over their physical locations. By providing fast yet physically structured intracranial field estimation, the proposed approach serves as a simulation backbone for practical tFUS digital twins. Our contributions are summarized as follows: (1) we formulate tFUS simulation as an operator learning problem—to our knowledge, the first operator-based formulation in this setting; (2) we develop a coordinate-aware neural operator that integrates free-field, skull image, transducer position and orientation in a shared coordinate frame, enabling a single model to handle data spanning multiple operating frequencies and transducer configurations; (3) we show that magnetic resonance (MR) input yields field predictions comparable to computed tomography (CT), supporting MR as a viable input for tFUS simulation.

\section{Methodology}

\subsection{Problem Formulation}\label{sec:2.1}
We formulate tFUS simulation as an operator learning problem \cite{JMLR:v24:21-1524} rather than a fixed voxel-to-voxel regression task. Let $\Omega_f \subset \mathbb{R}^3$ denote the focal region of interest (ROI) where the intracranial peak-pressure field is predicted, and let $\Omega_s \subset \mathbb{R}^3$ denote the skull ROI along the acoustic propagation path. The acoustic free-field is a function $p_{\mathrm{ff}}: \Omega_f \rightarrow \mathbb{R},$ describing the peak-pressure field generated by the transducer in a homogeneous medium without the skull, and the skull anatomy is a CT- or MR-derived function $h_{\mathrm{sk}}: \Omega_s \rightarrow \mathbb{R}$ serving as an anatomical proxy for the heterogeneous medium. Treatment parameters including transducer position and orientation are collected into a condition vector $\mathbf{c}$. The target is the intracranial peak-pressure field $p_{\max}(\mathbf{x}) = \max_t |p(\mathbf{x}, t)|$ on $\mathbf{x} \in \Omega_f$. Under this formulation, the skull-mediated acoustic propagation is represented by an operator that maps between function spaces. Let $\mathcal{F}(\Omega)$ denote the space of scalar functions on a domain $\Omega$, and $\mathcal{C}$ the space of condition vectors. The operator $\mathcal{G}$ can be written as
\begin{equation}
\mathcal{G}:\mathcal{F}(\Omega_f)\times\mathcal{F}(\Omega_s)\times\mathcal{C}\rightarrow\mathcal{F}(\Omega_f),
\quad
p_{\max}=\mathcal{G}(p_{\mathrm{ff}},h_{\mathrm{sk}},\mathbf{c}),
\end{equation}
which we approximate by a neural operator $\mathcal{G}_{\theta}$ parameterized by trainable weights $\theta$, i.e. $\hat{p}_{\max}=\mathcal{G}_{\theta}\left(p_{\mathrm{ff}},h_{\mathrm{sk}},\mathbf{c}\right).$

\subsection{Data Generation}
We follow the simulation framework of \cite{10.1007/978-3-032-07694-6_6} to generate paired free-field and intracranial fields and to incorporate skull images as an additional input modality. Following their pipeline, skull acoustic models are built from CT images of 13 subjects via Hounsfield unit (HU) segmentation, and a bowl-shaped single-element transducer is simulated with the k-Wave MATLAB toolbox \cite{10.1117/1.3360308}. We extend this framework in two respects: first, to cover a range of operating conditions, we simulate each skull at three frequencies (250, 400, and 500~kHz) and across 300 transducer placements per skull, yielding free-field and intracranial field pairs over diverse frequency conditions. Each simulated volume is cropped to a $56^3$ ROI centered on the target, at an isotropic spacing of 1 mm. Second, to represent both inputs in a shared physical coordinate frame, we assign each voxel a metric coordinate. For the focal and skull ROIs, we locate the center of each region, convert voxel indices to millimeter coordinates, and normalize them by the simulation domain size to $[-1, 1]^3$. The coordinate axes are defined with \textit{x}, \textit{y} and \textit{z} corresponding to the left–right, anterior–posterior, and superior–inferior directions of the skull, respectively.

\subsection{tFUSOperator}
Based on the formulation in Section \ref{sec:2.1}, we design \textit{tFUSOperator}, a coordinate-aware neural operator mapping free-field, skull anatomy, and treatment parameters to the intracranial field. Following recent neural operators that operate in a latent space for efficient computation \cite{alkin2025,wang2024,li2023}, we build \textit{tFUSOperator} as a Transformer \cite{vaswani2023} that maps the inputs to a fixed-size latent representation.

\subsubsection{Overview}
Let $\mathbf{X}_f=\{ \mathbf{x}^{f}_i \}_{i=1}^{N_f},$ and $\mathbf{X}_s=\{ \mathbf{x}^{s}_j \}_{j=1}^{N_s}$ be the voxel coordinates sampled on discrete grids in the focal and skull regions, where $N_f$ and $N_s$ are the numbers of sampled voxels in each region. The model receives the sampled value pairs $\{ ( \mathbf{x}^{f}_i, p_{\mathrm{ff}}(\mathbf{x}^{f}_i) ) \}_{i=1}^{N_f}$ and $\{ (\mathbf{x}^{s}_j, h_{\mathrm{sk}}(\mathbf{x}^{s}_j)) \}_{j=1}^{N_s},$ together with $\mathbf{c}$, and predicts $\hat{p}_{\max}$ on the focal domain. Given input pairs ${(\mathbf{X}_f, p_{\mathrm{ff}})}$ and ${(\mathbf{X}_s, h_{\mathrm{sk}})}$ with $\mathbf{c}$, the model can be written as
\begin{equation}
\hat{p}_{\max}=\mathcal{D}_{\theta}
(\mathcal{P}_{\theta}(\mathcal{E}_{\theta}(\mathbf{X}_f, p_{\mathrm{ff}}, \mathbf{X}_s, h_{\mathrm{sk}}), \mathcal{C}_{\theta}(\mathbf{c})
), \mathbf{X}_f),
\end{equation}
where $\mathcal{E}_{\theta}$, $\mathcal{C}_{\theta}$, $\mathcal{P}_{\theta}$, and $\mathcal{D}_{\theta}$ denote the encoder, conditioning module, latent processor, and decoder, respectively. The focal coordinates $\textbf{X}_f$ are reused as decoder queries so that the output is reconstructed on the same physical region where the free-field input is defined.

\subsubsection{Encoder}
The encoder maps the free-field and skull volumes into a fixed-size latent representation. Since $p_{\mathrm{ff}}$ and $h_{\mathrm{sk}}$ are defined on different physical regions, we use a dual-modality encoder with separate embedding branches but a shared coordinate frame. Each volume is tokenized into feature tokens ${\mathbf{a}^{f}_i}, {\mathbf{a}^{s}_j} \in \mathbb{R}^d$, where $d$ is the embedding dimension. The free-field volume is tokenized voxelwise by a $1 \times 1 \times 1$ convolution, whereas the skull volume is first passed through a local convolutional stem to better capture anatomical geometry before tokenization. Each token is paired with its normalized physical coordinate, embedded by a shared coordinate encoder
\begin{equation}
\mathbf{q}^{f}_i=\phi_{\mathrm{enc}}(\mathrm{PE}(\mathbf{x}^{f}_i)),\quad\mathbf{q}^{s}_j=\phi_{\mathrm{enc}}(\mathrm{PE}(\mathbf{x}^{s}_j)),
\end{equation}
where $\mathrm{PE}(\cdot)$ is a sinusoidal positional encoding and $\phi_{\mathrm{enc}}$ is shared across modalities. We then construct key and value tokens with a coordinate-decoupled attention design: keys generated from coordinate embeddings only, while values combine coordinate and feature tokens,
\begin{equation}
\begin{split}
&\mathbf{k}^{f}_i = \mathbf{q}^{f}_i,
\quad
\mathbf{v}^{f}_i=\phi_{\mathrm{val}}([\mathbf{q}^{f}_i ;\mathbf{a}^{f}_i])+\mathbf{e}_{f},\\
&\mathbf{k}^{s}_j = \mathbf{q}^{s}_j,
\quad
\mathbf{v}^{s}_j=\phi_{\mathrm{val}}
([\mathbf{q}^{s}_j ;\mathbf{a}^{s}_j])+\mathbf{e}_{s},
\end{split}
\end{equation}
where $\phi_{\mathrm{val}}$ is a shared value embedding network, $[\cdot;\cdot]$ denotes channel-wise concatenation, and $\mathbf{e}_{f}$ and $\mathbf{e}_{s}$ are learned modality embeddings added only to the values, leaving keys purely coordinate-dependent. Concatenating both modalities into $\mathbf{K} =
[\{{\mathbf{k}^{f}_i}\}_{i=1}^{N_f};{\{\mathbf{k}^{s}_j}\}_{j=1}^{N_s}]$ and $\mathbf{V} =[\{{\mathbf{v}^{f}_i}\}_{i=1}^{N_f};{\{\mathbf{v}^{s}_j}\}_{j=1}^{N_s}]$, a set of $M$ learnable latent queries $\mathbf{H}\in\mathbb{R}^{M\times d}$ attends to these coordinate-aware key-value tokens through cross-attention:
\begin{equation}
\mathbf{z}^0=\mathrm{CrossAttn}(\mathbf{H}, \mathbf{K}, \mathbf{V}),
\end{equation}
where $\mathbf{z}^0 \in \mathbb{R}^{M \times d}$ is the fixed-size latent representation passed to the latent processing module. This design separates \textit{where} the model attends, determined by coordinate-only keys, from \textit{what} information is read, represented by value tokens containing both spatial and modality-specific features.

\begin{figure*}[!ht]
    \includegraphics[width=\textwidth]{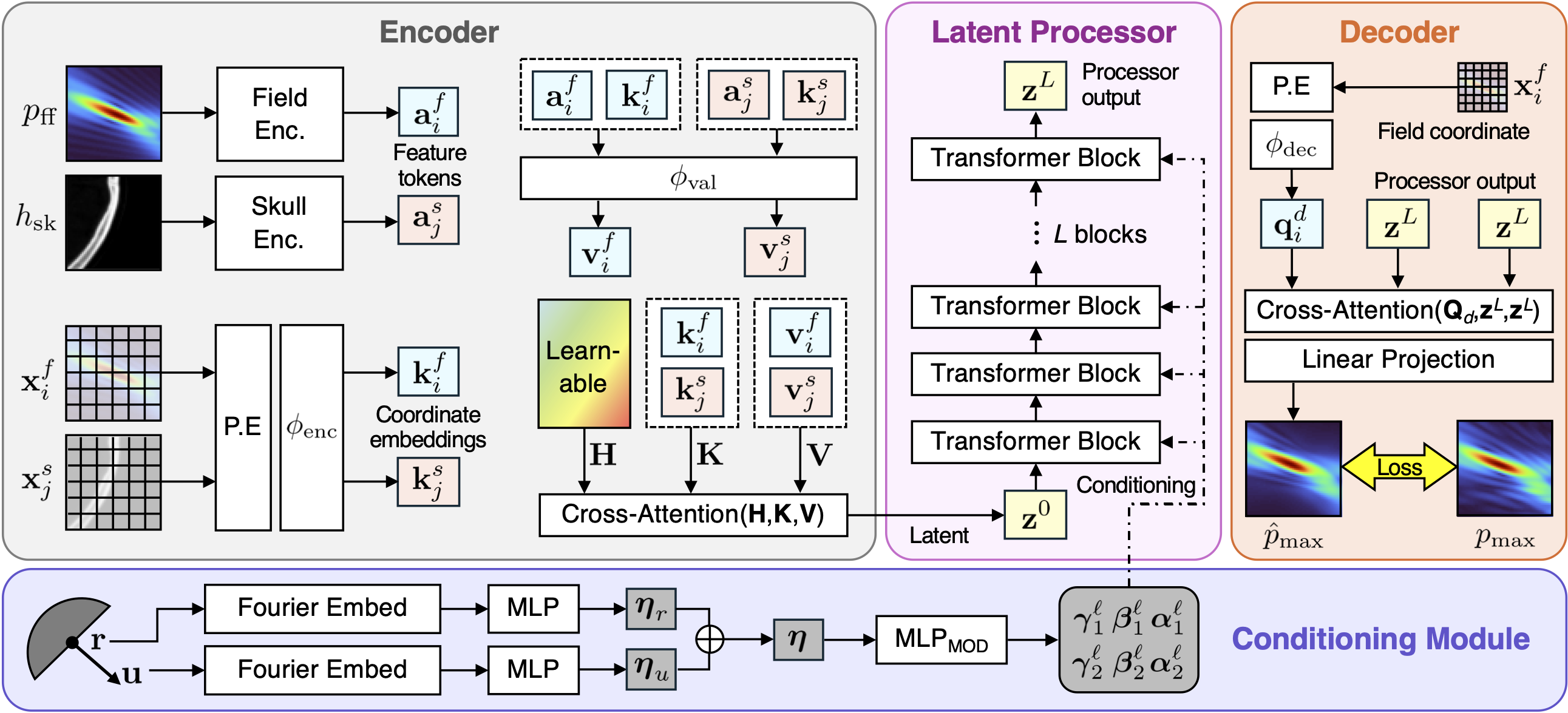}
\caption{Overview of \textit{tFUSOperator}. The model maps the free-field, skull volume, physical coordinates, and transducer parameters to the intracranial pressure field.} \label{fig:1}
\end{figure*}

\subsubsection{Conditioning Module}
The conditioning module embeds treatment parameters into a global condition vector for the latent processor. The module accepts an arbitrary set of scalar or vector parameters—such as transducer placement, geometry, or operating frequency—each encoded by a source-specific branch, so that additional parameters can be incorporated without architectural change. In this work, we instantiate two sources: transducer position $\mathbf{r} \in \mathbb{R}^{3}$, and focal-axis direction $\mathbf{u} \in \mathbb{R}^{3}$. For each source $\mathbf{s}\in\{\mathbf{r},\mathbf{u}\}$, the embedding is $\boldsymbol{\eta}_s=\psi_s(\gamma(\mathbf{s}))$ with Fourier feature mapping $\gamma(\cdot)$ and a source-specific multilayer perceptron (MLP) $\psi_s$, and the two are summed into $\boldsymbol{\eta}=\boldsymbol{\eta}_r+\boldsymbol{\eta}_u\in\mathbb{R}^d$. Each source embedding is randomly replaced by a learnable null embedding with probability 0.1 during training; at inference, this allows selected sources to be deactivated without changing the model architecture.

\subsubsection{Latent Processor}
The latent processing module propagates latent representation $\mathbf{z}^0$ produced by the encoder through $L$ transformer blocks ($\ell=0,\ldots, L-1$) conditioned on $\boldsymbol{\eta}$. Each block applies multi-head self-attention (MSA) followed by a feed-forward network (FFN). To inject transducer-dependent information, we use Diffusion Transformer (DiT)-style adaptive modulation \cite{peebles2023}: for each block, $\boldsymbol{\eta}$ is projected into six vectors $(\boldsymbol{\gamma}^{\ell}_{1},\boldsymbol{\beta}^{\ell}_{1},\boldsymbol{\alpha}^{\ell}_{1},\boldsymbol{\gamma}^{\ell}_{2},\boldsymbol{\beta}^{\ell}_{2},\boldsymbol{\alpha}^{\ell}_{2})=\mathrm{MLP}^{\ell}_{\mathrm{mod}}(\boldsymbol{\eta})$, where $\boldsymbol{\gamma}$, $\boldsymbol{\beta}$ are scale-and-shift and $\boldsymbol{\alpha}$ gates the residual. With $\mathrm{mod}(x,\boldsymbol{\gamma},\boldsymbol{\beta})
=(1+\boldsymbol{\gamma}) \odot x+\boldsymbol{\beta}$, each block is represented as
\begin{equation}
\begin{split}
&\tilde{\mathbf{z}}^{\ell}=\mathbf{z}^{\ell}+\boldsymbol{\alpha}^{\ell}_{1}
\odot
\mathrm{MSA}(\mathrm{mod}(\mathrm{LN}(\mathbf{z}^{\ell}),\boldsymbol{\gamma}^{\ell}_{1},\boldsymbol{\beta}^{\ell}_{1})),\\
&\mathbf{z}^{\ell+1}=\tilde{\mathbf{z}}^{\ell}+\boldsymbol{\alpha}^{\ell}_{2}
\odot
\mathrm{FFN}
(\mathrm{mod}(\mathrm{LN}(\tilde{\mathbf{z}}^{\ell}),\boldsymbol{\gamma}^{\ell}_{2},\boldsymbol{\beta}^{\ell}_{2})),
\end{split}
\end{equation}
where $\mathrm{LN}$ is layer normalization and $\odot$ is element-wise multiplication broadcast over the latent tokens. The modulation projection $\mathrm{MLP}_{\mathrm{mod}}$ is zero-initialized so that the conditioning pathway starts as an identity mapping ($\boldsymbol{\alpha}=0$) and gradually activates during training. The final output of the latent processor is denoted as $\mathbf{z}^{L}$ and is passed to the decoder.

\subsubsection{Decoder}
The decoder reconstructs $p_{\max}$ on $\Omega_f$ from the processed latent representation $\mathbf{z}^{L}$ using the focal coordinates $\mathbf{X}_f$ as queries. For each focal coordinate $\mathbf{x}^{f}_i \in \mathbf{X}_f$, a decoder query is formed as $\mathbf{q}^{d}_i
= \phi_{\mathrm{dec}}(\mathrm{PE}(\mathbf{x}^{f}_i)),$
where $\phi_{\mathrm{dec}}$ is a decoder-specific coordinate embedding network. The queries attend to the processed latent tokens through cross-attention:
\begin{equation}
\{\mathbf{a}^{d}_i \}_{i=1}^{N_f}=\mathrm{CrossAttn}(\mathbf{Q}_d, \mathbf{z}^{L}, \mathbf{z}^{L}),
\end{equation}
with $\mathbf{Q}_d={\{\mathbf{q}^{d}_i}\}$ and decoded focal tokens $\mathbf{a}^{d}_i\in\mathbb{R}^d$. A learned linear projection then maps each token to its scalar voxel value, and the values are arranged on the focal grid to give the output field $\hat{p}_{\max}: \Omega_f \rightarrow \mathbb{R}$. By querying the latent at output coordinates rather than decoding to a fixed grid, the decoder mirrors the operator formulation on the output side, reconstructing the field as a function evaluated at the focal coordinates where the free-field input is defined. An overall architecture of \textit{tFUSOperator} is illustrated in Fig.~\ref{fig:1}.

\section{Experiments}

\subsection{Implementation Details}

\subsubsection{Dataset}
Dataset was generated for 13 skulls and split at the skull level into 8 training, 2 validation, and 3 test subjects, ensuring that test skulls are entirely unseen during training. For each training skull, 30 of the 300 transducer positions were held out across all frequencies, providing a seen-skull, unseen-position evaluation set in addition to the unseen-skull test set. This separates two generalization regimes: interpolation to new transducer placements on known anatomy and transfer to unseen skull geometries.

\subsubsection{Model and Optimization}
The model was implemented in PyTorch 2.11.0 with CUDA 12.8 and trained on a single NVIDIA RTX A6000 GPU, using an embedding dimension of $d=384$, $M=512$ latent tokens, $L=8$ latent processor blocks with 6 attention heads, and a 3D residual skull stem with 3 residual blocks and 32 channels. The training objective combines voxel-wise mean squared error (MSE) reconstruction loss and soft Dice loss on the focal region, balanced by homoscedastic uncertainty weighting \cite{kendall2018} with a learnable per-term log-variance. We used AdamW ($\beta=(0.9, 0.95)$, weight decay $=0.05$), a peak learning rate of $2\times10^{-4}$, linear warmup, cosine decay, and a batch size of 4.

\subsubsection{Evaluation Metrics}
We evaluate the predicted intracranial peak-pressure field with three focus-oriented metrics. \textbf{Dice} measures the overlap between the full-width half-maximum (FWHM) regions of the predicted and target fields, reported as a percentage, capturing focal shape and location. \textbf{Peak distance ($\Delta_p$)} is the Euclidean distance (in mm) between the predicted and target peak-pressure locations, capturing focal positioning accuracy. \textbf{Peak difference ($\delta_p$)} is the relative error (in $\%$) between the predicted and target peak magnitudes, capturing focal peak-pressure magnitude accuracy.

\subsection{Results}
\begin{table}[t]
\centering
\caption{Quantitative results across evaluation regimes and input
modalities. Unseen pos denotes held-out transducer placements on training skulls (seen skull); unseen skull denotes test skulls excluded from training.}
\label{tab:1}
\begin{tabular}{llccc}
\hline
\textbf{Evaluation} & \textbf{Input} &
\textbf{Dice (\%)} $\uparrow$ &
$\Delta_{\mathrm{p}}$ \textbf{(mm)} $\downarrow$ &
$\delta_{\mathrm{p}}$ \textbf{(\%)} $\downarrow$ \\
\hline
\multirow{2}{*}{Unseen pos}
 & CT & 90.50 $\pm$ 5.94 & 1.31 $\pm$ 1.22 & 1.27 $\pm$ 1.43 \\
 & MR & 90.56 $\pm$ 6.37 & 1.33 $\pm$ 1.20 & 1.29 $\pm$ 1.24 \\
\hline
\multirow{2}{*}{Unseen skull}
 & CT & 72.41 $\pm$ 14.90 & 3.09 $\pm$ 2.23 & 1.80 $\pm$ 1.53 \\
 & MR & 71.54 $\pm$ 14.73 & 2.92 $\pm$ 2.19 & 1.95 $\pm$ 2.10 \\
\hline
\end{tabular}
\end{table}

Table~\ref{tab:1} reports performance across the two generalization regimes. On seen skulls with unseen transducer positions, the model predicts the intracranial focus with high fidelity, reaching about 90\% Dice with sub-1.4~mm peak localization. On entirely unseen skulls, performance decreases as expected, yet the model continues to localize the focus—around 72\% Dice with peak localization within roughly 3~mm—indicating that the operator transfers to new skull geometry despite the substantial anatomical variation across subjects. Notably, the peak-pressure magnitude is predicted accurately across all settings: the peak difference stays within about 2\% even on unseen skulls, where Dice and peak distance degrade more visibly.

\begin{figure*}[!ht]
\centering
    \begin{tabular}{c|c|c|c|c|c|c|c|c}
    \hline
    \multirow{2}{*}{\textbf{Freq}}& \multicolumn{4}{c|}{\textbf{Unseen pos}} & \multicolumn{4}{c}{\textbf{Unseen skull}}\\
    \cline{2-9}
     & $p_{\textrm{ff}}$ & $p_{\max}$ & $\hat{p}_{\max}^{\textrm{CT}}$ & $\hat{p}_{\max}^{\textrm{MR}}$ & $p_{\textrm{ff}}$ & $p_{\max}$ & $\hat{p}_{\max}^{\textrm{CT}}$ & $\hat{p}_{\max}^{\textrm{MR}}$\\
    \hline
    250 kHz
    & \figin{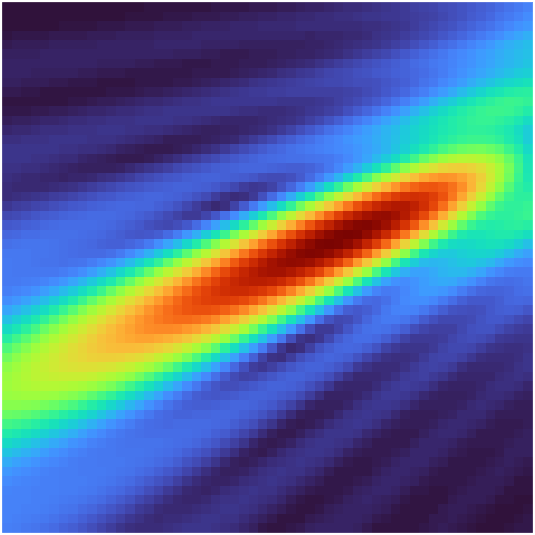}
    & \figin{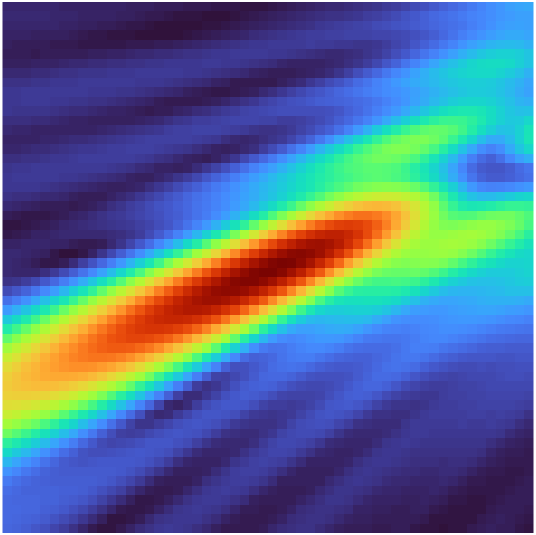}
    & \figin{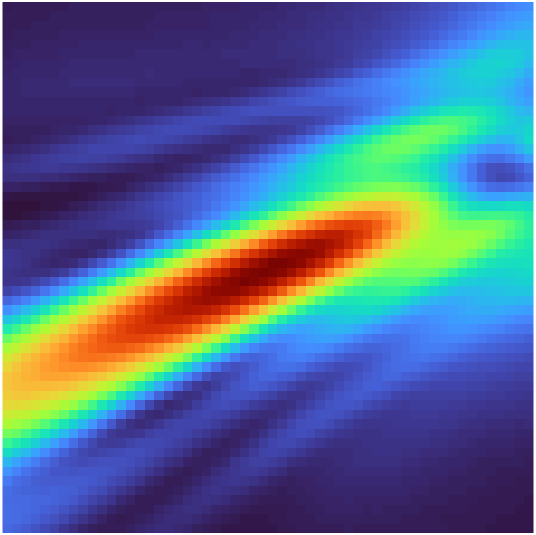}
    & \figin{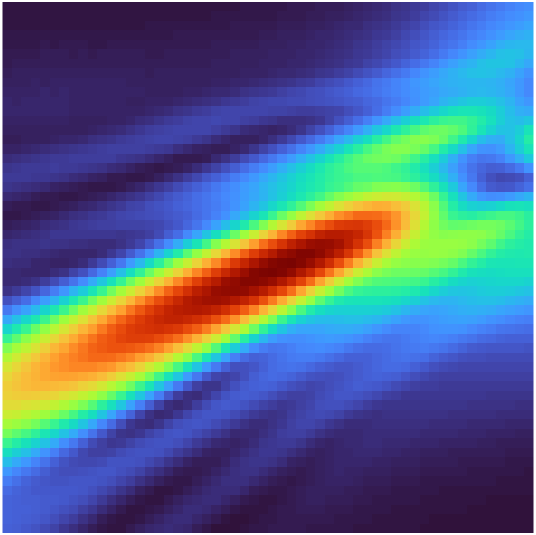}
    & \figin{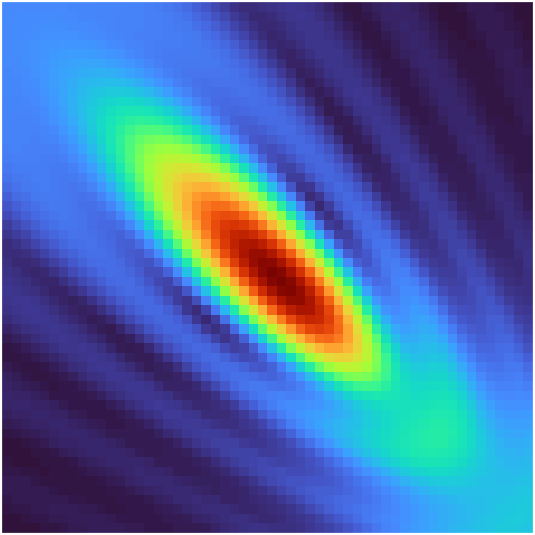}
    & \figin{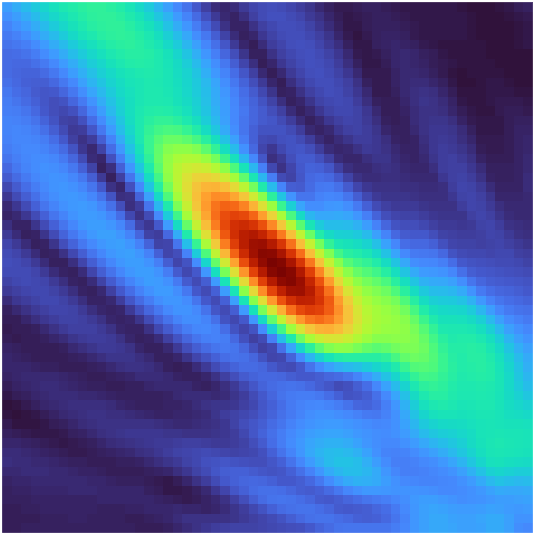}
    & \figin{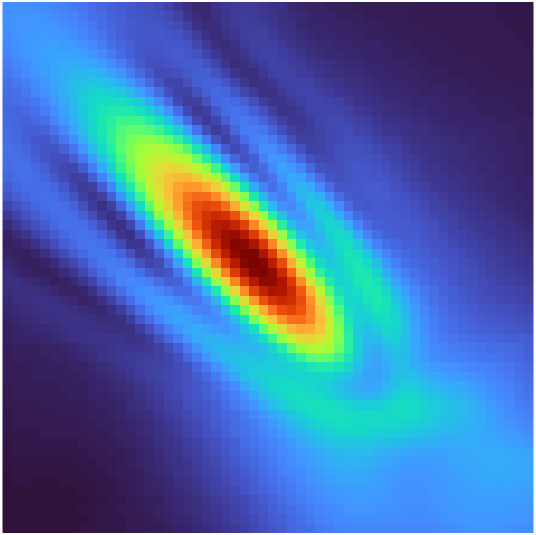}
    & \figin{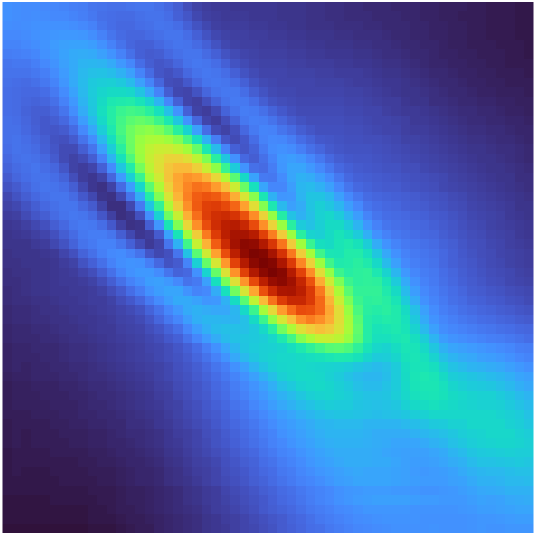}\\
    \hline
    400 kHz
    & \figin{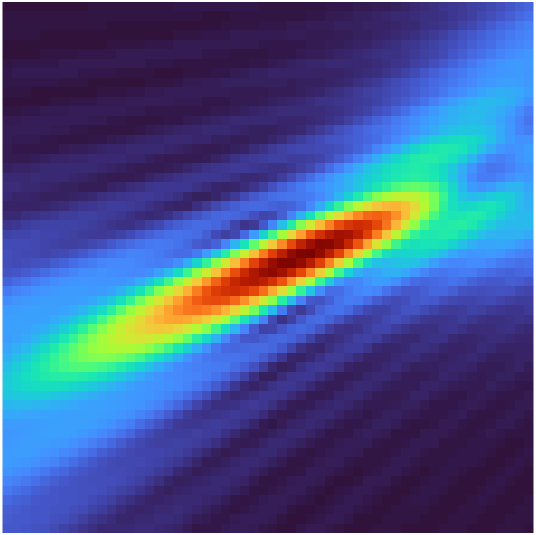}
    & \figin{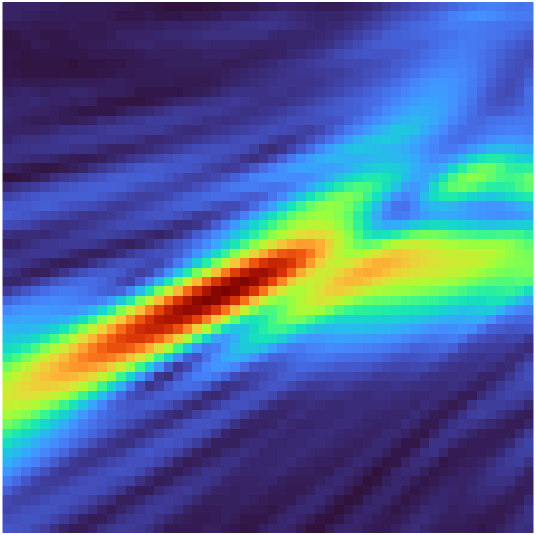}
    & \figin{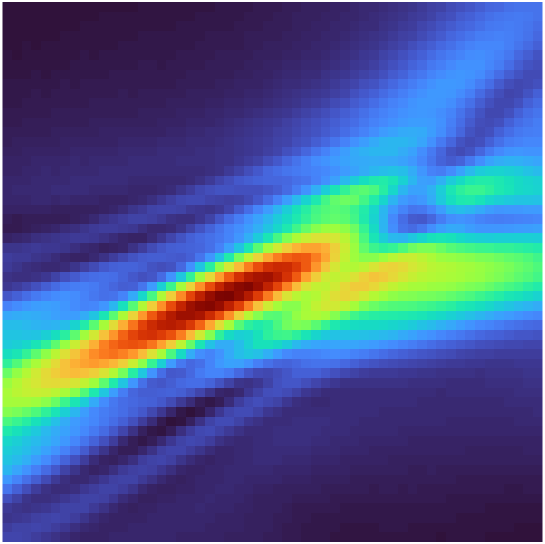}
    & \figin{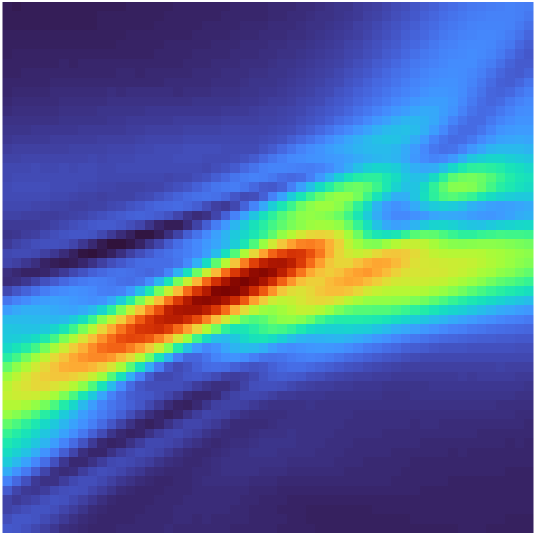}
    & \figin{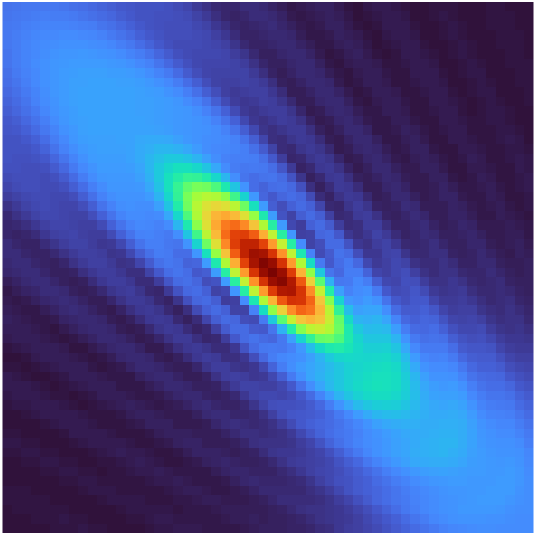}
    & \figin{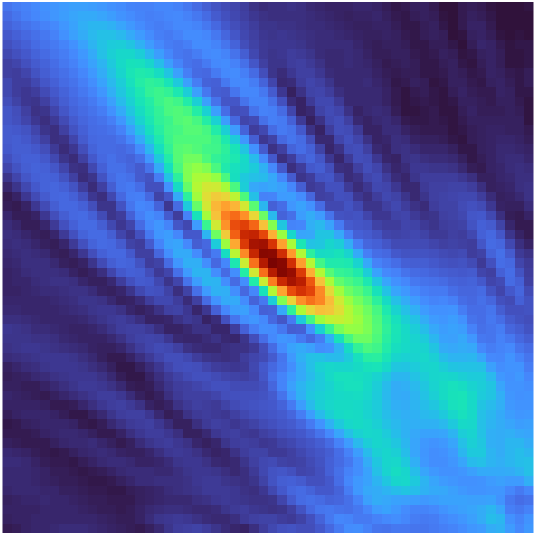}
    & \figin{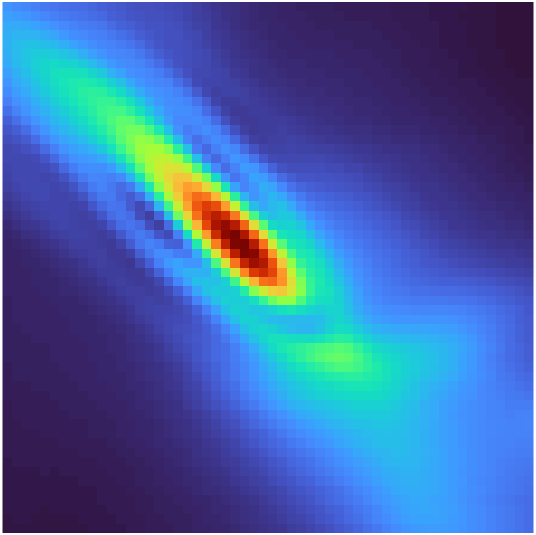}
    & \figin{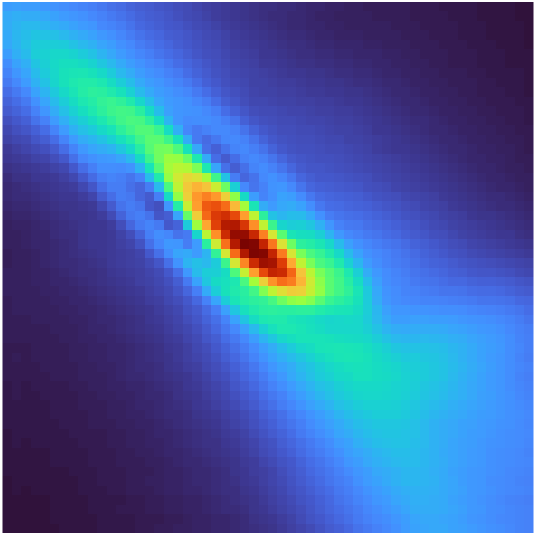}\\
    \hline
    500 kHz
    & \figin{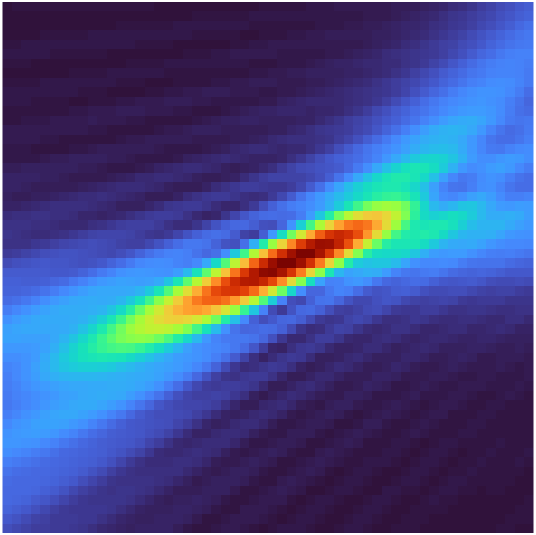}
    & \figin{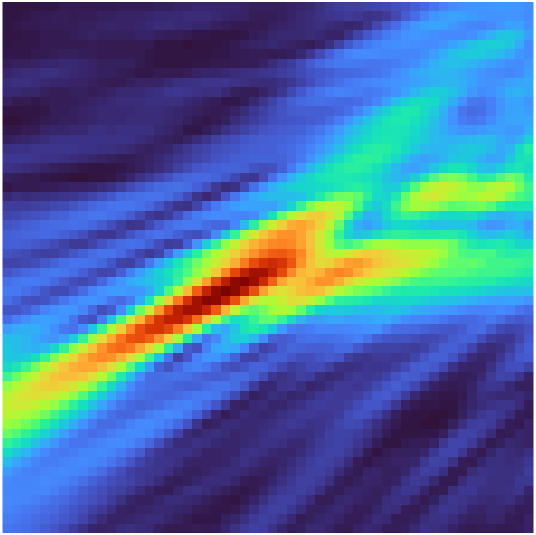}
    & \figin{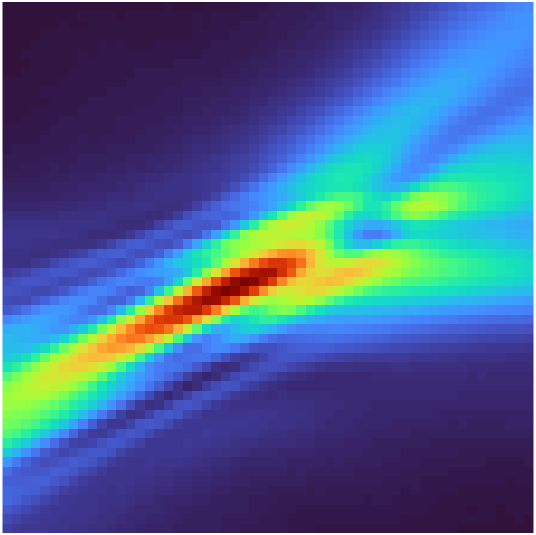}
    & \figin{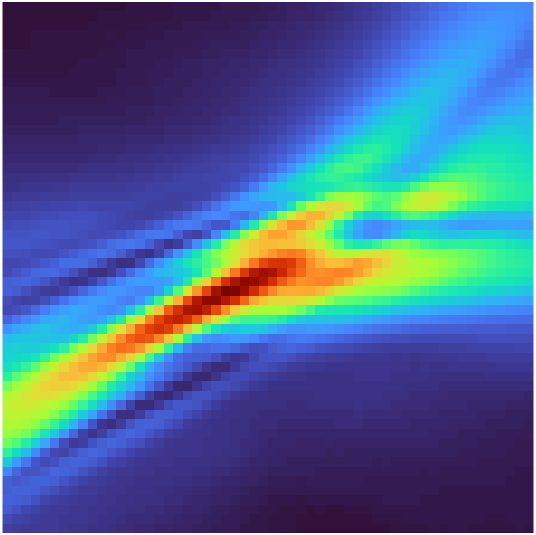}
    & \figin{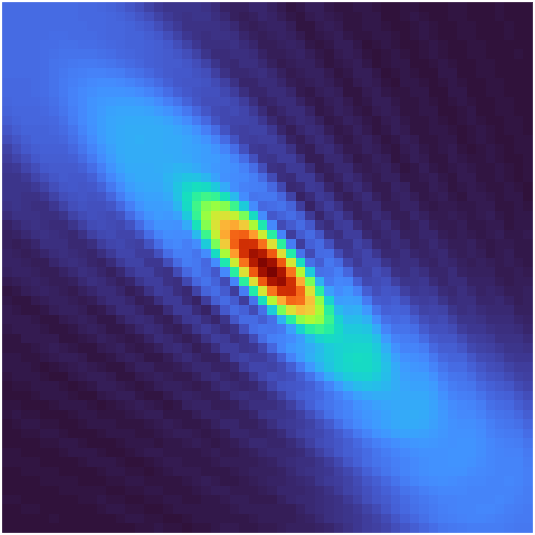}
    & \figin{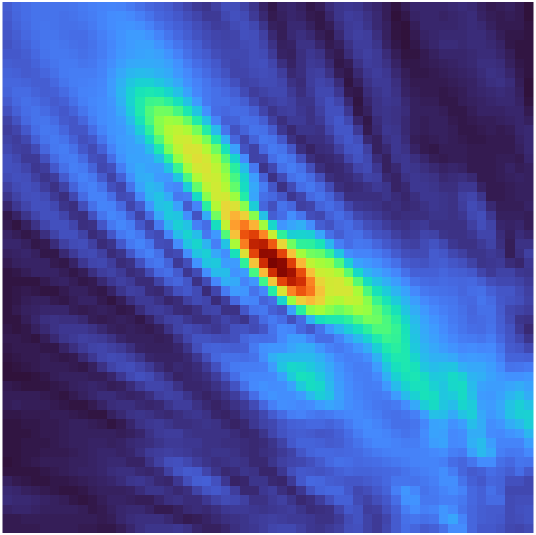}
    & \figin{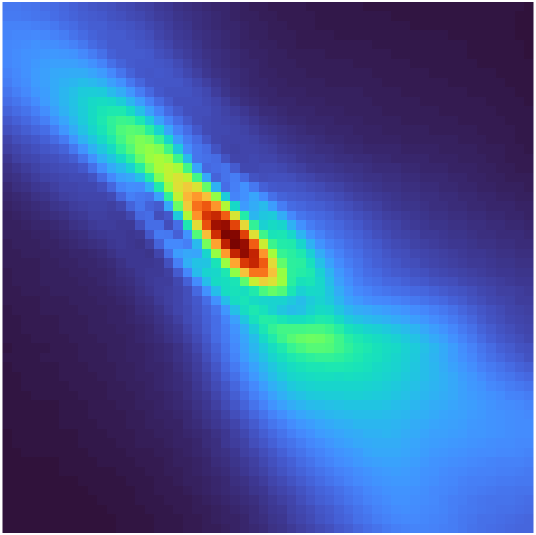}
    & \figin{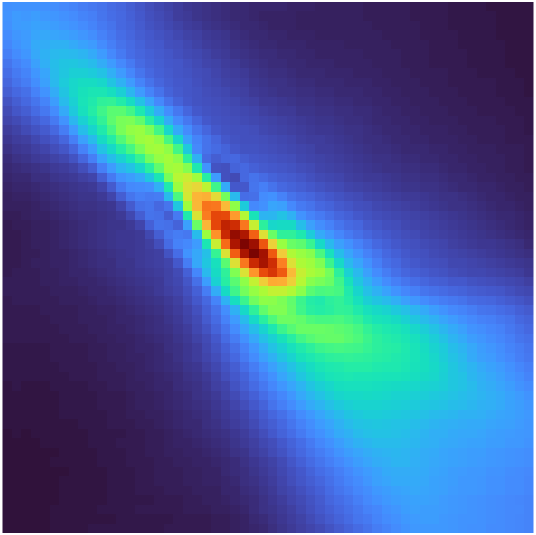}\\
    \hline
    \end{tabular}
    \caption{Visualization of free-field input ($p_{\textrm{ff}}$), target intracranial field ($p_{\max}$), and predictions from CT ($\hat{p}_{\max}^{\textrm{CT}}$) and MR ($\hat{p}_{\max}^{\textrm{MR}}$) input, shown on the \textit{xz}-plane through the focus for both evaluation regimes across three operating frequencies.}
    \label{fig:2}
\end{figure*}

Across both regimes, MR-based prediction is essentially on par with CT. The two modalities differ by less than one percentage point in Dice, and on unseen skulls MR matches CT in both peak distance and peak difference. This is notable because MR lacks the direct bone contrast that CT provides, yet the operator extracts comparable skull-aberration information from it. Figure~\ref{fig:2} shows that predictions closely follow the ground-truth focus in both shape and location. On seen skulls, CT and MR predictions are largely similar, but at higher frequencies the MR-based prediction reproduces the overall field noticeably closer to the ground truth than CT. On unseen skulls, the two modalities yield nearly identical predictions. Overall, these qualitative observations are consistent with the quantitative results, where MR performs on par with CT.

As a simulation backbone, the model is also orders of magnitude faster than full-wave solving: a single field prediction takes 2.16~ms on GPU, compared with 122~s for the k-Wave simulation on GPU, a 5.6$\times10^4$-fold speedup that makes the repeated, interactive field estimation required by a digital twin feasible.

\section{Discussion}
For a digital twin intended to support safe and efficient tFUS treatment, the choice of skull imaging modality is not only a matter of accuracy but also of clinical safety. CT clearly resolves bone structure and is the standard input for acoustic simulation, but acquiring it exposes the subject to ionizing radiation. A radiation-free, MR-derived simulation pathway is therefore highly desirable for safe and repeatable digital-twin operation. Against this backdrop, our finding that MR-based prediction nearly matches CT is a strong positive signal: the model can serve as a simulation backbone using MR alone, accurately recovering focal position and magnitude while reducing dependence on CT.

As a first step toward operator-based modeling of tFUS propagation, our study leaves room for further refinement. First, the evaluation is based on 13 subjects, and generalization to unseen skulls—the harder of the two regimes—is the main avenue for improvement. One promising direction is to pretrain the skull encoder on larger collections of skull images alone, so that it learns more transferable bone-geometry representations before being coupled to the operator. Second, the coordinate-query decoder predicts the field as a continuous function rather than on a fixed grid, and thus naturally supports prediction at arbitrary spatial resolutions. We have not yet exploited this property, which could enable resolution-flexible inference and finer focal-shape recovery—a direction that further leverages the operator formulation.

\section{Conclusion}
We presented \textit{tFUSOperator}, a coordinate-aware neural operator that casts transcranial ultrasound simulation as an operator learning problem, predicting the intracranial pressure field from the free-field input, skull anatomy, and treatment parameters within a shared physical coordinate frame. To our knowledge, this is the first operator-based formulation of tFUS field prediction, moving beyond fixed-grid regression. On both seen and unseen skulls the model localizes the acoustic focus accurately, and it does so from MR input nearly as well as from CT, while running several orders of magnitude faster than numerical full-wave simulation. Taken together, these results indicate that an operator-based surrogate can provide fast, radiation-free intracranial field estimation, an encouraging step toward safe and practical digital twins for patient-specific tFUS planning.

%
%
%
\bibliographystyle{splncs04}
\bibliography{references}

\end{document}